\documentclass[sigconf,nonacm]{acmart}
\AtBeginDocument{%
  \providecommand\BibTeX{{%
    \normalfont B\kern-0.5em{\scshape i\kern-0.25em b}\kern-0.8em\TeX}}}

\usepackage{multirow}
\usepackage{enumitem}

\renewcommand\footnotetextcopyrightpermission[1]{} 
\setlist{nolistsep}
\setcopyright{acmcopyright}
\copyrightyear{2018}
\acmYear{2018}
\acmDOI{XXXXXXX.XXXXXXX}

\begin{document}

\title{Common Knowledge Learning for Generating Transferable Adversarial Examples}

\author{Ruijie Yang}
\affiliation{%
  \institution{School of Computer Science and Engineering, Beihang University}
  \streetaddress{}
  \city{Beijing}
  \country{China}
}
\email{}

\author{Yuanfang Guo}
\affiliation{%
  \institution{School of Computer Science and Engineering, Beihang University}
  \streetaddress{}
  \city{Beijing}
  \country{China}
}
\email{}

\author{Junfu Wang}
\affiliation{%
  \institution{School of Computer Science and Engineering, Beihang University}
  \streetaddress{}
  \city{Beijing}
  \country{China}
}
\email{}

\author{Jiantao Zhou}
\affiliation{%
  \institution{State Key Laboratory of Internet of Things for Smart City \\ Department of Computer and Information Science, University of Macau}
  \streetaddress{}
  \city{Macau}
  \country{China}
}
\email{}

\author{Yunhong Wang}
\affiliation{%
  \institution{School of Computer Science and Engineering, Beihang University}
  \streetaddress{}
  \city{Beijing}
  \country{China}
}
\email{}

%
%
\renewcommand{\shortauthors}{}

\begin{abstract}
This paper focuses on an important type of black-box attacks, i.e., transfer-based adversarial attacks, where the adversary generates adversarial examples by a substitute (source) model and utilize them to attack an unseen target model, without knowing its information. Existing methods tend to give unsatisfactory adversarial transferability when the source and target models are from different types of DNN architectures (e.g. ResNet-18 and Swin Transformer). In this paper, we observe that the above phenomenon is induced by the output inconsistency problem. To alleviate this problem while effectively utilizing the existing DNN models, we propose a common knowledge learning (CKL) framework to learn better network weights to generate adversarial examples with better transferability, under fixed network architectures. Specifically, to reduce the model-specific features and obtain better output distributions, we construct a multi-teacher framework, where the knowledge is distilled from different teacher architectures into one student network. By considering that the gradient of input is usually utilized to generated adversarial examples, we impose constraints on the gradients between the student and teacher models, to further alleviate the output inconsistency problem and enhance the adversarial transferability. Extensive experiments demonstrate that our proposed work can significantly improve the adversarial transferability.
\end{abstract}





\maketitle
\begin{figure}[t]
  \centering
  \includegraphics[width=\linewidth]{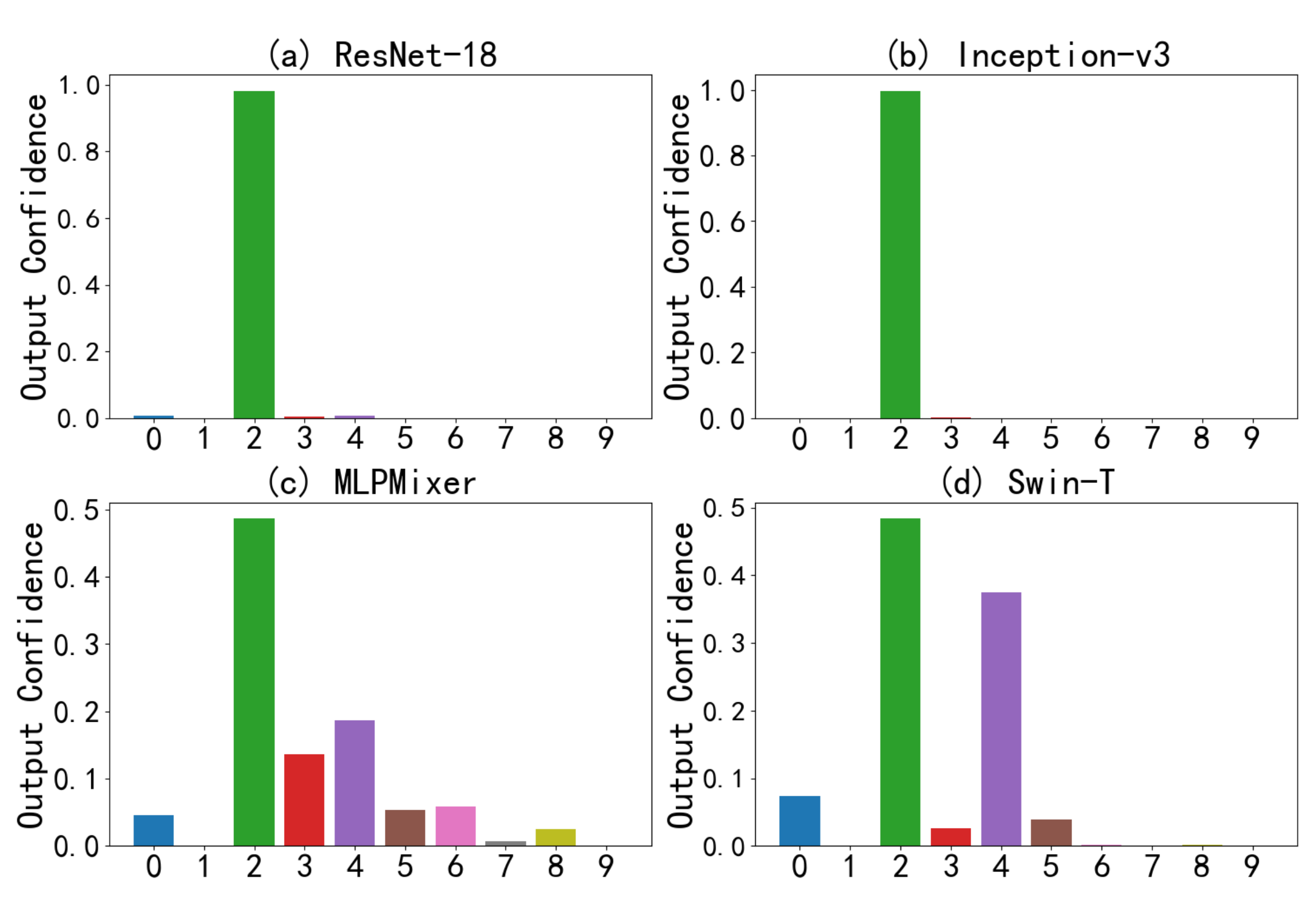}
  \vspace{-4mm}
  \caption{Output inconsistency among different networks with the same input image, whose truth label is class `2'. Although every model gives the correct prediction, the output probabilities are obviously different.}
  \vspace{-4mm}
  \label{fig:output_inconsistency}
\end{figure}

\begin{figure*}[t]
  \centering
  \includegraphics[width=1\linewidth]{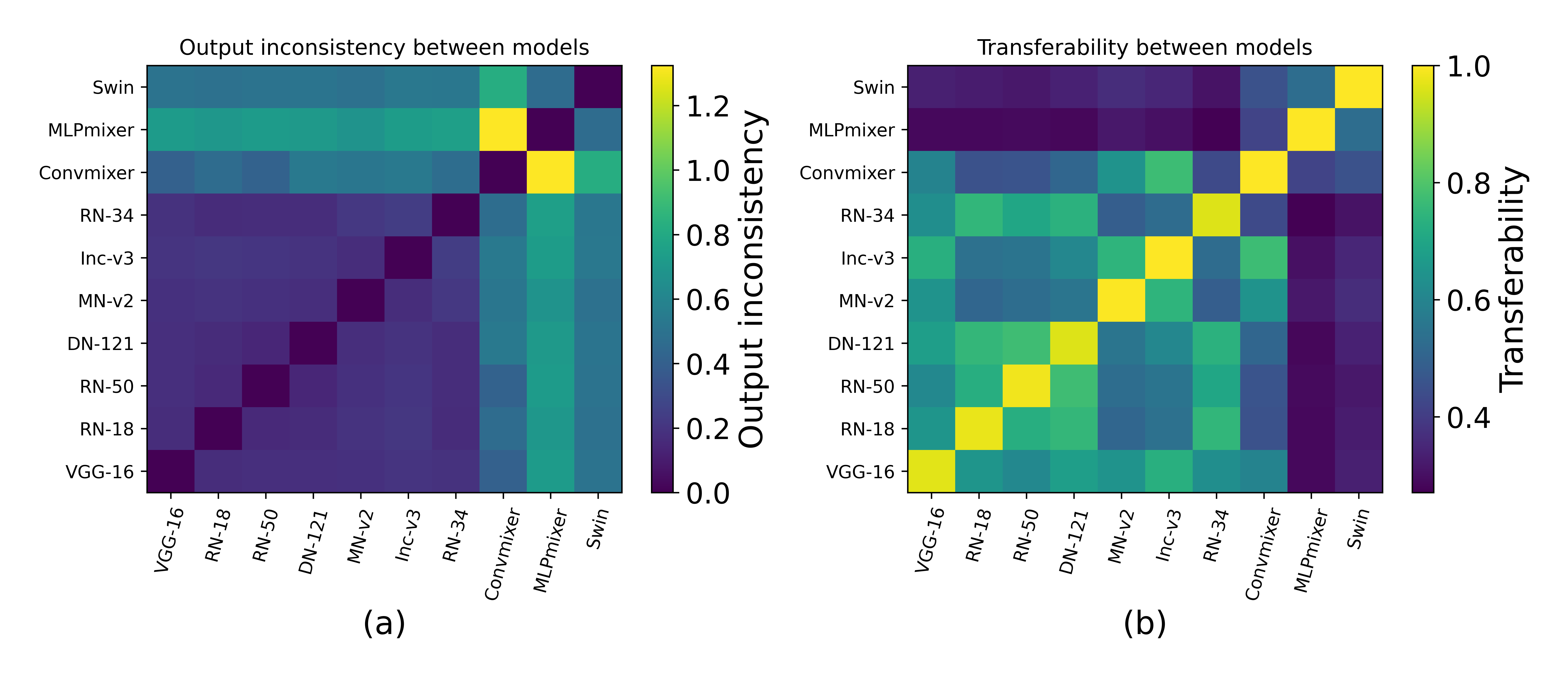}
  \vspace{-3mm}
  \caption{We calculate the output inconsistency and transferability on the CIFAR10 testing set. The results are obtained by averaging from 10,000 images. (a) The output inconsistency. A higher value denotes a higher inconsistency. (b) We take two models in turn as the source model to generate adversarial examples to attack the other and compute the transferability by averaging the two attack results. }
  \vspace{-1mm}
  \label{fig:heatmap}
\end{figure*}

\section{Introduction}

Recent studies have shown that deep neural networks (DNNs), such as convolutional neural networks (CNNs), transformers and etc., are vulnerable to adversarial attacks, which add special yet imperceptible perturbations to benign data to deceive the deep models to make wrong predictions. This vulnerability poses a considerable threat to the safety of the DNN-based systems, especially those applied in security-sensitive domains, such as autonomous driving, face-scan payment, etc.

Since different DNN architectures usually function differently, their corresponding vulnerabilities are also different. Therefore, existing adversarial attack techniques are usually specifically designed for different DNN architectures, to discover the safety threats of different DNN architectures.
Due to privacy or copyright protection considerations, black-box attacks tend to possess more applicability in real scenarios. In this paper, we focus on an extensively studied scenario of black-box attacks, i.e., transfer-based adversarial attack, which assumes that the adversary do not have access to the target model. Instead, the attacker can train substitute models to generate adversarial examples to attack the target model. For convenience, we only consider image classification DNN models in this paper.

For common CNN models, to enhance the transferability of adversarial examples, various techniques have been proposed, and they can be briefly classified into two categories according to their mechanisms, i.e., gradient modifications~\cite{mim,si-ni-fgsm,zou2022making} and input transformations~\cite{dim,tim,wang2021admix,ATTA}. The former type of methods usually improves the gradient ascend process for adversarial attacks to prevent the adversarial examples from over-fitting to the source model. The latter type of methods usually manipulates input images with various transformations, which enables the generated adversarial perturbations to adapt different input transformations. Consequently, these adversarial examples possess a higher probability of transferring to successfully attack the target unknown model.

The recent success of vision transformers (ViT) has also prompted several studies on devising successful attacks on the ViT type of architectures. ~\cite{DBLP:conf/iccv/MahmoodMD21,DBLP:conf/iclr/NaseerR0KP22,DBLP:conf/aaai/WeiCGWGJ22} construct substitute model based attack methods for ViTs, according to their unique architectures, such as attention module, classification token, to generate transferable adversarial examples.

Currently, existing methods usually employ pre-trained classification models as the source (substitute) model (as well as the target model in the experiments) directly, for achieving transfer-based adversarial attacks.
One of the core reasons, which affects the transferability of these adversarial attacks, is the similarity between the source (substitute) model and the target model.
By assuming that the source model and the target model are identical, the attack becomes a white-box attack. Then, the transferability is expected to be high, and the attack success rate is equivalent to that in the corresponding white-box setting. Intuitively, two models with similar architectures and similar weights tend to possess high transferability~\cite{waseda2023closer}.
On the contrary, models with significantly different architectures and weights usually exhibit low transferability.
For example, when we generate adversarial examples in ResNet-18 and test them on ViT-S, the attack success rate is 45.99\%, which is lower than the transferability from ResNet-18 to Inception-v3 (62.87\%).

Since different network architectures and weights will induce different outputs, we believe that the low adversarial transferability is due to the \textit{output inconsistency} problem, as depicted in Figure~\ref{fig:output_inconsistency}. As can be observed, even when each of the models gives correct classification result, the output probabilities are still inconsistent. Besides, the inconsistency between two different CNN models is usually smaller than that between two models from different architectural categories, e.g., a CNN model and a transformer-based model. Apparently, this inconsistency is harmful to adversarial transferability, because typical adversarial attacks are usually designed to manipulate the target model's output probability and this inconsistency will increase the uncertainty of the outputs of the target model.
To better describe this output inconsistency, KL divergence is employed to numerically represent it. As shown in Figure~\ref{fig:heatmap}, higher output inconsistencies tend to induce lower transferability and vice versa.

To alleviate the above problem, a straightforward solution is to construct a universal network architecture which possesses relatively similar output distributions as different types of DNN models. Unfortunately, this universal network architecture and its training strategy are both difficult to be designed and implemented. Besides, this solution is highly unlikely to effectively utilize the existing pre-defined DNN architectures, which are much more convenient to be applied in real scenarios.

To alleviate this \textit{output inconsistency} problem and effectively utilize the existing pre-defined DNN models, in this paper, we propose a common knowledge learning (CKL) method for the substitute (source) model to learn better network weights to generate adversarial examples with better transferability, under fixed network architectures. Specifically, to reduce the model-specific features and obtain better output distributions, we adopt a multi-teacher approach, where the knowledge is distilled from different teacher architectures into one student network. By considering that the gradient of input is usually utilized to generated adversarial examples, we impose constraints on the gradients between the student and teacher models. Since multiple teach models may generate conflicting gradients, which will interfere the optimization process, we adopt PCGrad~\cite{PCGrad} into our work to diminish the gradient conflicts of the teacher models.

Our contributions are summarized as follow.
\begin{itemize}
  \item We analyze the relationship between adversarial transferability and property of the substitute (source) model, and observe that a substitute model with less output inconsistency to the target model tends to possess better adversarial transferability.
  \item To reduce the model-specific features and obtain better output distributions, we propose a common knowledge learning framework to distill multi-teacher knowledge into one single student network.
  \item For generating adversarial examples with better transferability, we propose to learn the input gradients of the teacher models and utilize gradient projection to reduce the conflicts in the gradients of multiple teachers.
  \item Extensive experiments on CIFAR10 and CIFAR100 demonstrate that our method is effective and can be easily integrated into transfer-based adversarial attack methods to significantly improve their attack performances.
\end{itemize}

\section{Related Work}

\subsection{Adversarial Attacks}

Adversarial attack is firstly proposed by ~\cite{l-bfgs}. Subsequently, a large number of adversarial attack methods are proposed, which are usually classified into two categories according to the adversary's knowledge~\cite{realsafe}, i.e., white-box and black-box attacks. The black-box attacks can be further classified into query-based and transfer-based attacks. White-box attacks usually assume that the adversary can access all the necessary information, including the architecture, parameters and training strategy, of the target model. Query-based attacks usually assume that the adversary can obtain the outputs by querying the target model~\cite{nattack,hard-label-optimization,polish,zhou2022adversarial,lord2021attacking}. Transfer-based attacks generate adversarial examples without the access to the target model, whose assumption is the closest to practice. Under such circumstance, the adversary usually exploits a substitute model to generate adversarial examples and utilize the examples to deceive the target model~\cite{dim,tim,ila,mahmood2021robustness,yang2021enhancing}. Since our work focus on the transfer-based scenario, we will introduce this attack scenario in details in next subsection.

\subsection{Transfer-based Attacks}

Since different DNN architectures usually function differently, existing transfer-based attack techniques are usually specifically designed for different DNN architectures.

For CNN architectures, the attack approaches in transfer-based scenarios can mainly be classified into two categories, i.e., gradient modifications and input transformations. For gradient modifications based methods,~\cite{mim} firstly proposes MI-FGSM to stabilize the update directions with a momentum term to improve the transferability of adversarial examples.~\cite{si-ni-fgsm} adopts the Nesterov accelerated gradients into the iterative attacks.~\cite{zou2022making} proposes an Adam iterative fast gradient tanh method (AI-FGSM) to generate adversarial examples with high transferability. Besides,~\cite{yang2022adversarial} adopts the AdaBelief optimizer into the update of the gradients and constructs ABI-FGM to further boost the attack success rates of adversarial examples. Recently,~\cite{wang2021enhancing} introduces variance tuning to further enhance the adversarial transferability of iterative gradient-based attack methods.

On the contrary, input transformations based methods usually applies various transformations to the input image in each iteration to prevent the attack method from being overfitting to the substitute model.~\cite{tim} presents a translation-invariant attack method, named TIM, by optimizing a perturbation over an ensemble of translated images. Inspired by data augmentations,~\cite{yang2022adversarial} optimizes adversarial examples by adding image cropping operation to each iteration of the perturbation generation process. Recently, Admix~\cite{wang2021admix} calculates the gradient on the input image admixed with a small portion of each add-in image while using the original label, to craft adversarial examples with better transferability. ~\cite{ATTA} improves adversarial transferability via an adversarial transformation network, which studies efficient image transformations to boosting the transferability. ~\cite{yuan2021automa} proposes AutoMa to seek for a strong model augmentation policy based on reinforcement learning.

Since the above approaches are designed for CNNs, their performances degrade when their generated adversarial examples are directly input to other types of DNN architectures, such as vision transformers~\cite{vit}, mlpmixer~\cite{mlpmixer}, etc. Since the transformer-based architectures have also been widely applied in image classification task, several literatures have also presented transfer-based adversarial attack methods when transformer-based architectures are employed as the source (substitute) model.~\cite{DBLP:conf/iccv/MahmoodMD21} proposes a self-attention gradient attack (SAGA) to enhance the adversarial transferability.~\cite{DBLP:conf/iclr/NaseerR0KP22} introduces two novel strategies, i.e., self-ensemble and token refinement, to improve the adversarial transferability of vision transformers. Motivated by the observation that the gradients of attention in each head impair the generation of highly transferable adversarial examples, ~\cite{DBLP:conf/aaai/WeiCGWGJ22} presents a pay no attention (PNA) attack, which ignores the backpropagated gradient from the attention branch.

\subsection{Knowledge Distillation}

A common technique for transferring knowledge from one model to another is knowledge distillation. The mainstream knowledge distillation algorithms can be classified into three categories, i.e., response-based, feature-based and relation-based methods~\cite{gou2021knowledge}. The feature-based methods~\cite{passban2021alp,chen2021cross} exploits the outputs of intermediate layers in the teacher model to supervise the training of the student model. The relation-based method~\cite{lee2019graph} explores the relationships between different layers or data samples.
These two types of methods requires synchronized layers in both the teacher and student models. However, when the architectures of the teacher and student models are inconsistent, the selection of proper synchronized layers becomes difficult to achieve. On the contrary, ~\cite{hinton2015distilling}, which is a response-based method, constrains the logits layers of the teacher and student models, which can be easily implemented for different tasks without the above mentioned synchronization problem. Therefore, ~\cite{hinton2015distilling} is adopted in our proposed work.

\section{Methodology}

\subsection{Notations}
Here, we define the notations which will be utilized in the rest of this paper. Let $x\in \mathcal{X}\subseteq R^{C\times W \times H}$ denote a clean image and $y$ is its corresponding label, where $\mathcal{X}$ is the image space. Let $z=(o_1, o_2, ..., o_i, ..., o_K)\in R^K$ be the output logit of a DNN model, where $K$ is the number of classes. $T_1(\cdot), T_2(\cdot), ...,T_n(\cdot)$ are employed to denote the teacher networks. $S(\cdot)$ stands for the student network. Correspondingly, $L_S(\cdot)$, $L_{T_i}(\cdot)$ are utilized to denote the losses of the student and teacher models, respectively. The goal of an generated adversarial example $x^*$ is to deceive the target DNN model, such that the prediction result of the target model $F_{tar}(\cdot)$ is not $y$, i.e., $\underset{i}{argmax} \ F_{tar}(x^*)\neq y$. Meanwhile, the adversarial example is usually desired to be similar to the original input, which is usually achieved by constraining the adversarial perturbation by $l_{p}$ norm, i.e., $\|x^*-x\|_p \le \epsilon$, where $\epsilon$ is a predefined small constant.

\begin{figure*}[t]
  \centering
  \includegraphics[width=\linewidth]{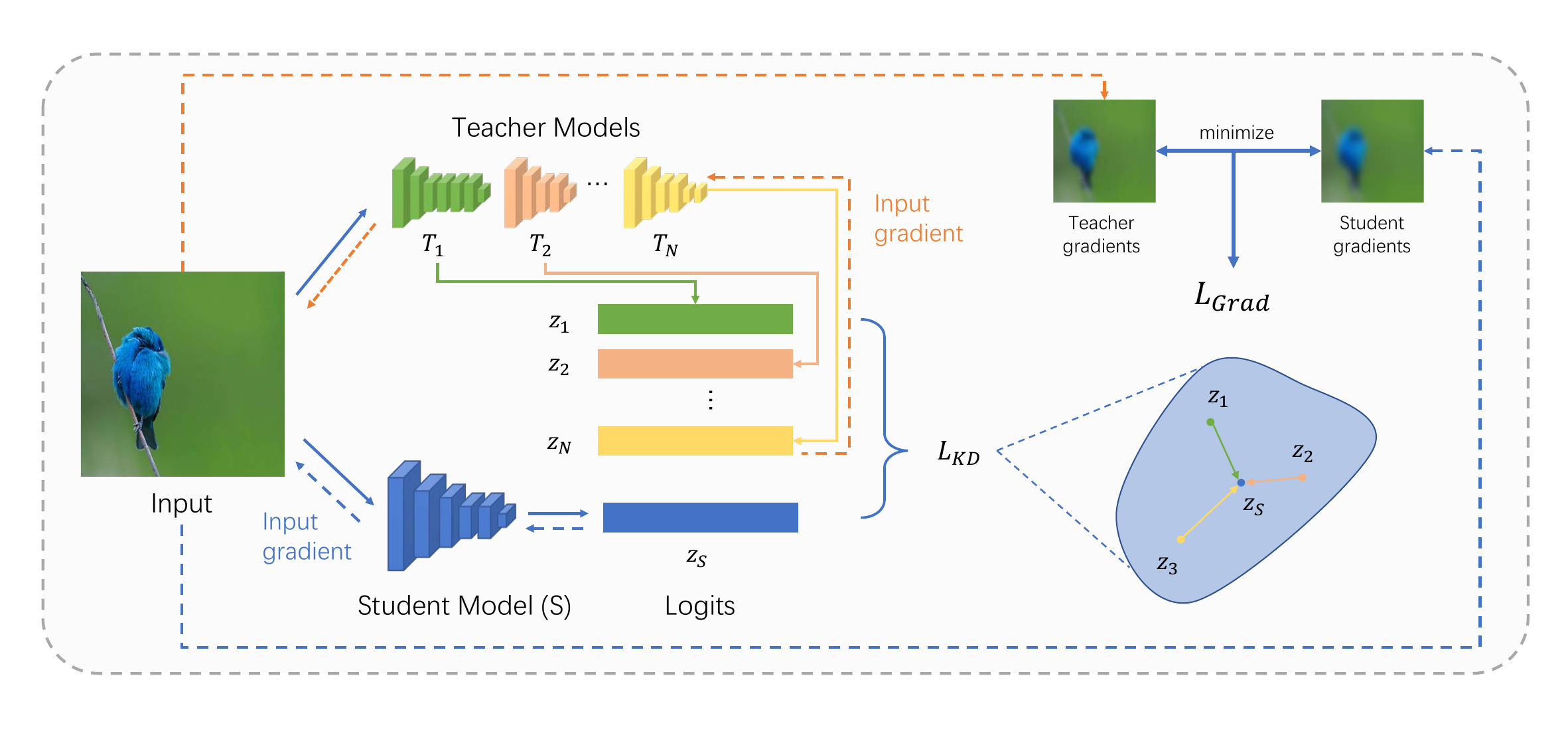}
  \vspace{-8mm}
  \caption{Our common knowledge learning (CKL) framework. We leverage the input gradient distillation loss and knowledge distillation loss to force the student model (S) to learn the common knowledge from multiple teacher models.}
  \vspace{-4mm}
  \label{fig:framework}
\end{figure*}

\subsection{Overview}

According to Figures \ref{fig:output_inconsistency} and \ref{fig:heatmap}, we can observe that the output inconsistency problem significantly affects the transferability of adversarial examples, i.e., high output inconsistency usually indicates low transferability, and vice versa. Since the output inconsistency within each type of DNN architectures is usually less than that of the cross-architecture models, the adversarial examples generated based on the substitute model from one type of DNN architectures (e.g. CNNs) usually give relatively poor attack performance on the target models from other types of DNN architectures (e.g. ViT, MLPMixer). The straightforward solution to alleviate the output inconsistency problem, i.e., designing a new universal network architecture and its training strategy, are quite difficult, inefficient and inconvenient for real scenarios.

To alleviate the aforementioned problems, in this paper, we propose a common knowledge learning (CKL) framework, which distills the knowledge of multiple teacher models with different architectures into a single student model, to obtain better substitute models. The overall framework is shown in Figure~\ref{fig:framework}. Firstly, we select teacher models from different types of DNN architectures. The student model will learn from their outputs to reduce the model-specific features and obtain common (model-agnostic) features, to alleviate the output inconsistency problem. Since the input gradient is always utilized in typical adversarial attack process, the student model will also learn from the input gradients of the teacher models to further promote the transferability of the generated adversarial examples.
After training the student model, in the test stage, this student model will be utilized as the source (substitute) model to generate adversarial examples.

\subsection{Common Knowledge Distillation}

As can be observed from Figure~\ref{fig:output_inconsistency}, when the same input is fed into different models, the output probabilities are quite different, which actually reveals that there exists feature preference in deep models. Apparently, the output inconsistency problem is induced by these model-specific features, because these features, which are considered to be distinctive to one model, may not be distinctive enough to others. Under such circumstance, when an adversarial example is misclassified by the source model to the other class $\hat{y}$($\hat{y} \neq y$), it contains certain manipulated features which are distinctive to the source model. However, if these manipulated features are not distinctive enough to the target model, the adversarial example may not be able to  deceive the target model.

According to the analysis above, it is crucial to identify and emphasize the common (model-agnostic) features among different DNN models in the substitute model, such that when these model-agnostic features is manipulated to generate the adversarial examples, the target model will possess a higher possibility to be deceived, i.e., the adversarial transferability will be improved. Therefore, we construct a multi-teachers knowledge distillation method to force the student model to learn and emphasize the common features from various DNN models. Since different DNN models usually possess different architectures, we constrain the model outputs between the student and teacher models, by adopting ~\cite{hinton2015distilling}. Specifically, the KL divergence is exploited to measure the output discrepancy between each teacher model $T_i(\cdot)$ and the student model $S(\cdot)$, which is formulated as
\begin{equation}\label{eq:kl-div}
  KLdiv(T_i(x), S(x)) = \sum_{k=1}^{K} S(x)_k \cdot log \frac{S(x)_k}{T_i(x)_k},
\end{equation}
where $K$ represents the number of classes.

To jointly utilize all the teacher models, the knowledge distillation (KD) loss $L_{KD}$ is defined as
\begin{equation}\label{eq:kd_loss}
  L_{KD} = \Sigma_{i=1}^N KLdiv(T_i(x), S(x)).
\end{equation}

\subsection{Gradient Distillation}

Since the input gradients are commonly utilized in the main-stream adversarial attack methods, such as FGSM, MIM, DIM, TIM, VNI-FGSM etc., if two networks $F(x)$ and $G(x)$ satisfy $\nabla_x L_F(x)=\nabla_x L_G(x)$, adversarial examples generated by these methods will be identical when either of the two networks are employed as the source model. Additionally, if $\nabla_x L_F(x)=\nabla_x L_G(x)$, the losses of $F(x)$ and $G(x)$ differ by at most one constant, i.e., there exists a constant $C$ that $L_F(x) = L_G(x) + C$. Since the outputs of two models are more likely to be inconsistent when their losses are different, if the input gradients of two models are similar, these two models tend to possess less output inconsistency. Although this assumption cannot be exactly satisfied in real scenarios, it can still be useful in generating transferable adversarial examples. Therefore, we constrain the student model to learn the input gradients from the teacher networks, to further improve the adversarial transferability.

Since our framework will utilize multiple teacher networks to teach one student model, it is vital to design suitable approaches to learn multiple input gradients. Under such circumstance, a simple solution is to design a multi-objective optimization problem which minimizes the distances between the input gradients of the student model and each teacher model.
This optimization problem can be simplified by minimizing the distance between the input gradient of the student model and the averaged input gradients of the teacher models (which can be regarded as a representative value among all the input gradients of the teacher models), as
\begin{equation}\label{mean_grad}
  \min \|\nabla_x L_S(x) - \bar{g}(x)\|_2^2,
\end{equation}
where $\bar{g}(x)= \Sigma_{i=1}^N g_i(x)$ and $g_i(x) = \nabla_x L_{T_i}(x)$. For convenience, we let $g_i$ denote $g_i(x)$ and let $\bar{g}$ denote $\bar{g}(x)$ in the rest of this section, when the input $x$ is not necessary to be emphasized.

Unfortunately, there exists conflicting gradients among them, which is similar to the gradient conflict problem in multi-task learning~\cite{cagrad}. This gradient conflict problem means that there exists a $j$, satisfying $g_j \cdot \bar{g} < 0$. If $\nabla_x L_S(x)$ is gradually closer to $\bar{g}$, this gradient conflict problem will actually move $\nabla_x L_S(x)$ further away from $g_j$, which is the input gradient of the $j$-th teacher model.

To address this issue, inspired by the PCGrad~\cite{PCGrad} method, we adjust our optimization objective function by projecting one of the two conflicting gradients onto the normal plane of the other gradient.
Specifically, when we have two conflicting gradients, $g_i$ and $g_j$, we will replace $g_i$ with $g_i = g_i - \frac{g_i\cdot g_j}{g_j\cdot g_j} g_j$. This replacement step is performed for all the gradients. After the replacement step, we calculate $d(x)=\Sigma_{i=1}^N g_i(x)$ as the ground truth for the student model to learn. Then, the gradient objective function becomes
\begin{equation}\label{eq:grad_loss}
  L_{Grad} = \|\nabla_x L_S(x) - d(x)\|_2^2.
\end{equation}
By combining Eq.~\ref{eq:kd_loss} and Eq.~\ref{eq:grad_loss}, the final loss of our common knowledge learning (CKL) for training the student network can be obtained as
\begin{equation}\label{eq:total_loss}
  L = L_{KD} + \lambda \cdot L_{Grad},
\end{equation}
where $\lambda$ is the hyperparameter employed to balance the two loss terms.

\subsection{Generating Adversarial Examples with CKL}

After the training process, we utilize the trained student model (S) as the source (substitute) model to generate adversarial examples.
Our framework can be easily combined with the existing transfer-based adversarial attack methods. For demonstration, here we leverage DI-FGSM~\cite{dim} as an example to explain the adversarial example generation process. Let $CE(\cdot)$ denote the commonly utilized Cross Entropy loss. Let $\varphi (\cdot)$ denote the input transformations, i.e., resizing and padding~\cite{dim}. We set $x_0 = x$. Then, the adversarial example generation process can be formulated as
\begin{equation}
\begin{aligned}
  L(x_t) & = CE(S(\varphi (x_t)), y) \\
  g_t & = \nabla_{x_t} L(x_t) \\
  m_{t+1} & = \mu m_t + \frac{g_t}{\|g_t\|_1} \\
  x_{t+1} & = Clip_{x, \epsilon}(x_t + \alpha \cdot sign(m_{t+1})),
\end{aligned}
\end{equation}
where $\epsilon$ is a predefined small constant to constrain the maximum magnitudes of the generated adversarial example. $Clip_{x, \epsilon}(\cdot)$ forces the modified value to stay inside the $L_\infty$ ball ($\{x_t | \|x_t-x\|_\infty \le \epsilon \}$). $\alpha$ is the step size and $\mu$ is momentum. This process terminates when $t$ reaches the maximum number (N) of iterations, and $x_{N}$ is the finally generated adversarial example.

\begin{table*}
  \caption{Non-targeted attack results on CIFAR10. The first column introduces the source models and the first row presents the target models. We report the averaged attack success rate on the entire testing set. `*' denotes the teacher models. $^{\bigtriangleup}$ implies that the source and target model s are identical. MI-FGSM, DI-FGSM and VNI-FGSM are abbreviated as `MI', `DI', `VNI', respectively. `+CKL' represents that our CKL framework is integrated.}
  \label{tab:cifar10_results}
  \begin{tabular}{c|c|cccc|cccc}
    \toprule
                                 &Attack Method & ResNet-50*   & Inception-v3* & Swin-T*  & MLPMixer*        & VGG-16  & DenseNet-121 & ConvMixer & ViT-S  \\
    \midrule
    \multirow{6}{*} {ResNet-18}  &MI            & 70.67        & 62.81        & 45.99      & 39.17         &73.73    & 74.75        & 58.22     & 37.17  \\
                                 &MI+CKL        & 86.30        & 83.09         & 82.02      & 71.78         & 87.66   & 88.37        & 81.71     & 62.14 \\
                                 &DI            & 77.84        & 69.39         & 56.17      & 44.80         & 79.77   & 83.10        & 65.68     & 42.73\\
                                 &DI+CKL        & 93.17        & 89.75         & 90.07      & 81.57         & 93.77   & 94.67        & 89.62     & 73.56 \\
                                 &VNI           & 76.43        & 71.94         & 53.68     & 43.22         & 80.05     &80.36          &67.46      & 41.05  \\
                                 &VNI+CKL       & 88.90       & 86.64         & 86.95     & 78.55         & 90.45     &90.80          &86.81      & 68.82  \\
    \midrule
        \multirow{6}{*} {VGG-16} &MI            & 49.97        & 69.80         & 43.55      & 36.11         & $96.84^{\bigtriangleup}$   & 57.69         & 64.81    & 32.17 \\
                                 &MI+CKL        & 79.46         & 91.89        & 84.60      & 68.19         & 95.74   & 84.08         & 91.54     & 58.40 \\
                                 &DI            & 59.65         & 79.62        & 53.97      & 41.13         & $99.76^{\bigtriangleup}$     &71.26        & 73.19       & 37.76 \\
                                 &DI+CKL        & 88.10         & 96.42        & 91.68      & 78.97         & 98.56     &92.30        & 95.92       & 70.85\\
                                 &VNI           & 52.27         & 79.95        & 54.10      & 40.67         & $99.73^{\bigtriangleup}$     &61.98      & 75.43         &34.88  \\
                                 &VNI+CKL       & 83.55         & 96.66        & 91.69      & 78.42         & 98.71     &89.14      & 96.73         & 67.63  \\
    \midrule
        \multirow{6}{*} {Swin-T} & MI             & 16.39         & 25.45        & $100^{\bigtriangleup}$        & 48.17         & 23.49     &17.23      & 43.61         & 38.43 \\
                                 &MI+CKL        & 26.25         & 39.02        & 99.79      & 78.16         & 37.28     &27.64      & 59.12         & 55.10\\
                                 &DI            & 26.43         & 36.28        & $100^{\bigtriangleup}$        & 57.93         & 35.57     &28.57      & 56.94         & 50.86\\
                                 &DI+CKL        & 36.87         & 51.84         &99.93      & 83.75         & 50.09     &40.72      & 72.16         & 66.31  \\
                                 &VNI           & 16.29      & 27.04          & $100^{\bigtriangleup}$         & 51.84         & 24.58     &16.65      & 48.26        & 39.71   \\
                                 &VNI+CKL       & 27.37      & 43.69            &99.92      & 84.75         & 40.75     &29.23      & 66.80        & 60.40   \\
        \midrule
        \multirow{6}{*} {ViT-S} & MI            & 19.05        & 24.70        & 51.20   & 69.74             & 23.15   & 18.80       &42.90      &$100^{\bigtriangleup}$ \\
                                 &MI+CKL        & 44.66        & 56.58        & 82.66   & 86.86             & 56.43   & 47.32       &68.75      &83.58 \\
                                 &DI            & 23.14        & 29.11        & 56.80   & 72.19             & 26.80   & 23.26       &50.20      &$99.98^{\bigtriangleup}$\\
                                 &DI+CKL        & 54.69        & 67.26        & 88.98   & 90.88             & 66.18   & 58.12       &78.54      &89.45  \\
                                 &VNI           & 20.60        & 26.71        & 45.80   & 73.56             & 25.13   & 20.44       &46.46      &$100^{\bigtriangleup}$  \\
                                 &VNI+CKL       & 47.40        & 48.64        & 86.45   & 89.69             & 59.47   & 49.87       &73.27      &86.13     \\
    \bottomrule
  \end{tabular}
  \vspace{-4mm}
\end{table*}

\section{Experiments}

In this section, we firstly introduce the necessary information for our experiments. Then, we present the non-targeted results in Sec.~\ref{sec:non-target-c10} and Sec.~\ref{sec:non-target-c100}. Next, in Sec.~\ref{sec:target_attack}, we conduct the experiments in targeted attack scenario. At last, we conduct an ablation study on the effects of our proposed modules in Sec.~\ref{sec:ablation}.

\subsection{Experimental Settings}
\label{Sec:exp_setting}

\noindent \textbf{Datasets.} Two widely used classification datasets, i.e., CIFAR10 and CIFAR100, are employed in our experiments. Both of two datasets possess images with the size of $32\times 32 \times 3$. For each dataset, 50,000 images are selected as the training set for training the student model, and 10,000 images are selected as the testing set for generating adversarial examples.

\noindent \textbf{Networks.} Nine networks with different types of DNN architectures are employed as either source model or target model, which includes ResNets~\cite{resnet}, VGG-16~\cite{vgg}, DenseNet-121~\cite{densenet}, Inception-v3~\cite{inceptionv3}, MobileNet-v2~\cite{mobilenetv2}, ViT-S~\cite{vit}, Swin Transformers~\cite{swin}, MLPMixer~\cite{mlpmixer}, and ConvMixer~\cite{convmixer}. To learn common knowledge from different types of DNN architectures, ResNet-50, Inception-v3, Swin-T, and MLPMixer are constructed as the teacher models.

\noindent \textbf{Baselines.} Our method is compared with several attacks, including MI-FGSM~\cite{mim}, DI-FGSM~\cite{dim}, VNI-FGSM~\cite{wang2021enhancing}, and ILA-DA~\cite{ila-da}. The numbers of attack iterations $M$ is set to 30 and step size is set to $1/255$ in all the experiments.

\noindent \textbf{Implementation Details.} We employ the training set to train the student model (S) and testing set for generating the adversarial examples. In the training process, the momentum SGD optimizer is employed, with an initial learning rate $lr=0.1$ (annealed down to zero following a cosine schedule), momentum $0.9$, and weight decay $0.0003$. The maximum epoch number is $600$. In the attack stage, we constrain the adversarial example and origin input by the $l_\infty$ ball with $\epsilon$ = 8/255, i.e., $\|x^*-x\|_\infty \le 8/255$. For DI-FGSM~\cite{dim}, each input benign image is randomly resized to $rnd \times rnd \times 3$, with $rnd \in [28, 32)$, and then padded to the size $32 \times 32 \times 3$ in a random manner.

\noindent \textbf{Evaluation Metric.} The attack success rate (ASR) is employed to evaluate the attack performance. It is defined as the probability that the target model is fooled by the generated adversarial examples, i.e.,
\begin{equation}\label{eq:asr-formula}
  ASR = 1 - \frac{\#\{correct \ samples\}}{\#\{total \ samples\}},
\end{equation}
where $\#$ denotes the number of elements in the set.

\subsection{Non-targeted Attack Results on CIFAR10}
\label{sec:non-target-c10}

The attack success rates (ASR) of the non-targeted attack on CIFAR10 are reported in Table~\ref{tab:cifar10_results}. Note that the elements in the first row and column represent the target and source models, respectively. We compare our method with MI-FGSM, DI-FGSM, and VNI-FGSM, which are abbreviated as MI, DI, and NI, respectively. Results comparison to ILA-DA~\cite{ila-da} are provided in the supplementary material due to the space limit. Their original methods generate adversarial examples by directly employing the pre-trained models. Meanwhile, our adversarial examples are generated by the student model, which is trained by our CKL framework. As can be observed, our CKL can give significant improvements compared to the corresponding baseline methods, which proves the effectiveness of our proposed work for adversarial transferability.

\begin{table*}
  \caption{Non-targeted attack results on CIFAR100. The first column introduces the source models and the first row presents the target models. The second column gives the attack methods. We report the averaged attack success rate on the entire testing set. DI-FGSM and VNI-FGSM are abbreviated as `DI' and `VNI', respectively. `+CKL' represents that our CKL framework is integrated.}
  \label{tab:c100}
  \begin{tabular}{c|c|cccccc|c}
    \toprule
                                 &Attack Method           & VGG-16    & DenseNet-121    & MLPMixer  & ConvMixer &Swin-S & Swin-B    & Average \\
    \midrule
    \multirow{4}{*} {ResNet-18}      &DI                & 88.97     & 85.62     & 65.16     & 72.79     &73.69  &73.48      & 76.62\\
                                 &DI+CKL            & 95.05     & 91.47     & 82.47     & 84.72     &83.03  &82.96      & \pmb{86.62 (+10.0)}\\
                                 &VNI               & 89.69     & 85.48     & 64.45     & 79.32     &73.01  &71.89      & 77.31\\
                                 &VNI+CKL           & 95.03     & 90.90     & 82.14     & 84.25     &82.82  &82.57      & \pmb{86.29 (+8.98)}\\
    \midrule
    \multirow{4}{*} {ResNet-50}      &DI                & 86.90     & 85.76     & 66.85     & 76.55     &74.68  &73.89      & 77.44    \\
                                 &DI+CKL            & 94.47     & 92.44     & 85.04     & 86.98     &84.75  &84.27      & \pmb{87.99 (+10.55)}\\
                                 &VNI               & 87.46     & 84.27     & 66.33     & 77.15     &74.85  &73.90      & 77.33\\
                                 &VNI+CKL           & 94.99     & 92.07     & 84.70     & 87.47     &85.30  &84.78      & \pmb{88.22 (+10.89)}\\
    \midrule
    \multirow{4}{*} {Swin-T}     &DI                & 57.87     & 48.88     & 79.24     & 65.27     &83.12  &81.40      & 69.30\\
                                 &DI+CKL            & 68.90     & 59.22     & 83.48     & 72.79     &88.18  &86.37      & \pmb{76.49 (+7.19)}\\
                                 &VNI               & 52.78     & 41.10     & 73.53     & 61.94     &81.81  &79.23      & 65.07\\
                                 &VNI+CKL           & 62.76     & 51.26     & 82.74     & 68.88     &86.85  &83.64      & \pmb{72.69 (+7.62)}\\
  \bottomrule
\end{tabular}
\vspace{-4mm}
\end{table*}

\noindent \textbf{Transferability to the unseen models.} Note that ResNet-50, Inception-v3, Swin Transformer, and, MLPMixer are employed as teacher models and utilized to train the student models. As can be observed, when selected these four models as the target models, the results with our CKL framework are significantly improved, compared to their corresponding baselines. To better verify the effectiveness of our CLK, we also employ the unseen models, e.g., VGG-16, DenseNet-121, and ViT-S, for evaluations, and our CKL can also achieve significant improvements. For example, when the source model is ResNet-18 and the target model is ViT-S, our method can obtain up to 25\% gains. This phenomenon indicates that our CKL framework can learn effective common knowledge from the teacher models and leverage them to the unseen models.

\noindent \textbf{Transferability to the cross-architecture models.} The cross-architecture transferability is usually a challenging problem for the baseline attack methods, as can be observed from the results. For example, when the source model is selected as ViT-S, the correspondingly generated adversarial examples' transferability to DenseNet-121 is relatively low, i.e., only 23.26\% for DI-FGSM. On the contrary, by integrating our CKL framework, the attack transferability can be largely improved, due to the common knowledge learning at the training stage.

\subsection{Non-targeted Attack Results on CIFAR100}
\label{sec:non-target-c100}
For better assessment of our proposed work, we further validate our CKL method on CIFAR100 and the results are shown in Table~\ref{tab:c100}. The experimental setups are identical to these in Section~\ref{Sec:exp_setting}. DI-FGSM and VNI-FGSM are employed as the baseline methods.

As can be observed, our method has a consistent improvement on the CIFAR100 dataset, whatever the attack method and the source model are. Besides, as shown in the last column of Table~\ref{tab:c100}, which reports the averaged ASRs of the test models, our CKL method can improve the averaged value for at least 7 percentage points, compared to the corresponding baseline methods. In addition, for the cross-architecture transferability, our method usually gives an improvement of more than 10 percentage points. For example, when the source model is ResNet-50 and the target model is MLPMixer, `DI+CKL' outperforms `DI' up to 18.19\% and `VNI+CKL' outperforms `VNI' up to 18.37\%. These results further verify the effectiveness of our CKL method.

\begin{table}
  \caption{Targeted attack results on CIFAR10. The first column introduces the source models and the first row presents the target models. Targeted VNI-FGSM and Logit attack are selected as the baseline methods.}
  \label{tab:target}
  \begin{tabular}{c|ccccc}
    \toprule
                                 &Attack            & VGG-16    & DN-121    & ConvMixer & ViT-S\\
    \midrule
    \multirow{4}{*} {RN-18}      &VNI              & 45.51     & 48.00     &27.01      & 8.05 \\
                                 &VNI+CKL          & 70.79       & 71.39     &59.46      & 29.23 \\
                                 &Logit            &  48.81     & 58.36     &26.58      & 9.91\\
                                 &Logit+CKL        &  72.34     & 77.45     &54.42      & 33.70\\
    \midrule
    \multirow{4}{*} {Swin-T}       &VNI            &  4.26      & 2.79      &10.96      & 6.89\\
                                 &VNI+CKL          &  10.85     & 6.23      &23.16      & 17.92\\
                                 &Logit            & 15.62      & 10.91     &24.16      & 17.93\\
                                 &Logit+CKL        & 31.42      & 23.79     &45.86      & 33.73 \\
  \bottomrule
\end{tabular}
\vspace{-4mm}
\end{table}

\subsection{Targeted Attack Results on CIFAR10}
\label{sec:target_attack}

Here, we evaluate the effectiveness of our proposed method on targeted adversarial attack. The targeted attack requires the target model to classify the adversarial examples into a pre-specific class $t(t\neq y)$, while the non-targeted attack only requires the model to make a wrong prediction. Thus, the targeted attack is indeed a more challenging problem. Note that the targeted attack results on CIFAR100 are provided in the supplementary material due to limited space.

To evaluate the performance of our method, we employ two baseline methods, i.e., VNIFGSM~\cite{wang2021enhancing} and Logit attack~\cite{logit_attack}. By following~\cite{logit_attack}, we set the maximum number of iterations to 300, step size to 2/255 and $\epsilon$ to 8/255. We randomly generate a target label $t(t\neq y)$ for each data pair $(x, y)$. We utilize the testing set to generate adversarial examples and employ ResNet-18 and Swin Transformer as the source models. The generated adversarial examples are tested on VGG-16, DenseNet-121, Convmixer, and ViT-S, which are not overlapped with the teacher models. Note that for targeted attack, the attack success rate $tASR$ is computed as
\begin{equation}\label{eq:tasr}
  tASR = \frac{\#\{x\in \mathcal{X}' | \underset{i}{argmax}F_{tar}(x)=t \}}{\#\{x\in \mathcal{X}'\}},
\end{equation}
where $F_{tar}(x)$ denotes the output class of the target model and $\mathcal{X}'$ represents the adversarial examples set.

As can be observed, the tASR scores, as shown in Table~\ref{tab:target}, are usually significantly lower than the corresponding ASR scores, as shown in Table~\ref{tab:cifar10_results}, with the same settings. Besides, we can observe that the cross-architecture transfer attack usually leads to lower tASR values, compared to attacking a target model, which is in the same type of DNN architectures. For example, when VNI is employed as the attack method and ResNet-18 is utilized as its source model, ConvMixer only obtains 27.01\% tASR score, because the distinctive features of ResNet-18 and ConvMixer tend to be different. On the contrary, when our CKL framework is integrated, the corresponding results obtain large gains, e.g., up to 22.45\% improvement for the above example. This phenomenon also indicates that our CKL framework can enable the student (substitute) model to learn common knowledge from multiple teacher models, which significantly improves the adversarial transferability.

\subsection{Ablation Study}
\label{sec:ablation}

\noindent \textbf{Input Gradient Distillation Scheme.}
Here, the effectiveness of our input gradient distillation scheme is validated. For better comparisons, we firstly introduce several variants of the objective function in learning the input gradients.

(i) The student model is training without gradient learning, i.e., it only employs Eq.~\ref{eq:kd_loss} as the objective function, which is denoted as `w/o teacher gradients'.

(ii) The objective function is replaced by the averaged value of multiple teacher models' gradients, i.e., $\|\nabla_x L_S(x) - \bar{g}(x)\|_2^2$, where $\bar{g}(x)= \Sigma_{i=1}^N g_i(x)$, which is denoted as `average of teacher gradients'.

(iii) The gradient objective function is replaced by the maximum input gradient value of the teacher models. Considering that the positive and negative signs of the gradients do not affect the final results, we select the max absolute value of gradients. This objective function is set to $\|\nabla_x L_S(x) - \bar{g}(x)\|_2^2$, where $g_{max}(x)=g_i(x)$ and $|g_i(x)|=\underset{j}{\max}|g_j(x)|$. This variant is denoted as `max of teacher gradients'.

Here, ResNet-18 is employed as the source model. The teacher models are identical to that in Section~\ref{Sec:exp_setting}. As can be observed in Table~\ref{tab:grad_ablation}, learning the input gradient from teacher models is effective. Moreover, our method is more effective than these variants.

\begin{figure}[t]
  \centering
  \includegraphics[width=\linewidth]{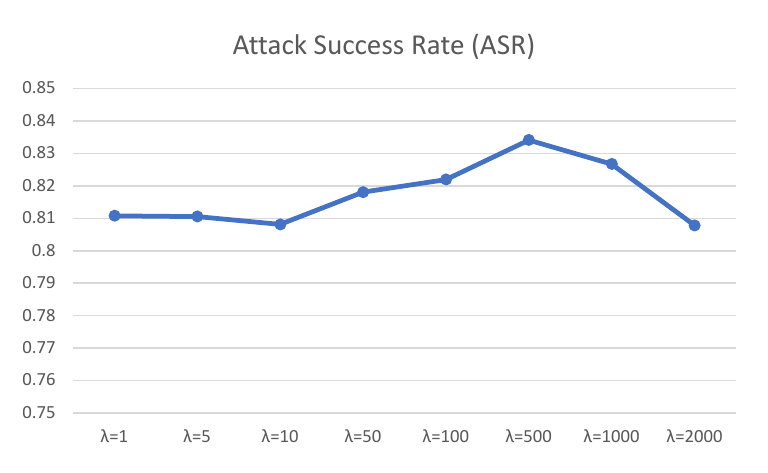}
  \caption{Attack results with different $\lambda$ values. The x-axis is the $\lambda$ value and y-axis denotes the attack success rate. The source (student) model is ResNet-18 and ASR is calculated as the averaged ASR values of nine target models. }
  \label{fig:lambda_alter}
  \vspace{-4mm}
\end{figure}

\noindent \textbf{Effects of $\lambda$.}
Here, we study the effects of the hyperparameter $\lambda$ in Eq.~\ref{eq:total_loss}. To assess its impacts, we set $\lambda=$ 1, 5, 10, 50, 100, 500, 1000, 2000. ResNet-18 is employed as the source model and The target models include VGG-16, ResNet-18, ResNet-50, DenseNet-121, Inception-v3, ConvMixer, MLPMixer, Swin-T, and ViT-S. The results are shown in Figure~\ref{fig:lambda_alter}. As can be observed, when $\lambda$ value is small, the objective function in Eq.~\ref{eq:total_loss} is dominanted by the first term, i.e., the objective function of knowledge distillation, and the result remains essentially unchanged. With the increasing of $\lambda$, the second term, i.e., the objective function of input gradient distillation, starts to function gradually, and thus the performance gradually increases. However, when $\lambda$ is relatively large, e.g., $\lambda=1000$, the attack success rate will decline. Thus, to achieve a good balance between the knowlege distillation and input gradient distillation objectives on all the models, we select $\lambda=500$ in our experiments.

\begin{table}
  \caption{Ablation study on different objective functions for input gradient distillation. The source (student) model is ResNet-18.}
  \label{tab:grad_ablation}
  \begin{tabular}{cccc}
    \toprule
                                      &VGG-16         & Swin-T       & ViT-S \\
    \midrule
    w/o teacher gradients             & 86.90         & 78.85     & 59.93 \\
    average of teacher gradients        & 87.65         & 81.36     & 61.50 \\
    max of teacher gradients            & 87.39         & 81.92     & 61.94 \\
    \midrule
    ours                           & \pmb{87.66}  & \pmb{82.02}  & \pmb{62.14} \\
  \bottomrule
\end{tabular}
\vspace{-4mm}
\end{table}

\begin{table}
  \caption{Ablation Study on the selection of different teacher models. The source (student) model is ResNet-18. }
  \label{tab:teacher_ablation}
  \begin{tabular}{c|ccccc}
    \toprule
                                & RN-18        & RN-50    & DN-121    & ConvMixer  &Swin-T \\
    \midrule
    teacher (a)                 & 83.93       & 79.72    & 81.87           & 67.90     &  51.13        \\
    teacher (b)                 & \pmb{89.51}       & \pmb{86.30}    & \pmb{88.37}         & 81.71     & 82.02 \\
    teacher (c)                 & 83.69       & 78.80    & 82.80          & \pmb{89.80}     & \pmb{91.40}        \\
  \bottomrule
\end{tabular}
\vspace{-4mm}
\end{table}

\noindent \textbf{Selection of the Teacher Models.}
To learn a common knowledge from different types of DNN architectures, ResNet-50, Inception-v3, Swin-T, and MLPMixer are employed as the teacher models in our experiments. Here, we study the effects of different selections of the teacher models, i.e.,

(a). 4 CNN models, including ResNet-18, ResNet-50, Inception-v3, and VGG-16;

(b). 2 CNN models and 2 non-CNN models, including ResNet-50, Inception-v3, Swin-T, and MLPMixer;

(c). 4 non-CNN models, including Swin-T, MLPMixer, ConvMixer, and ViT-S.

For convenience, ResNet-18 is employed as the student (source) model. The results are shown in Table~\ref{tab:teacher_ablation}, where RN-18, RN-50 and DN-121 are the abbreviations of ResNet-18, ResNet-50 and DenseNet-121, respectively. As can be observed, when the teacher models are all selected from the non-CNN models, the adversarial transferability to the non-CNN models is relatively high while that to the CNN models is relative low, because the student model learns more bias from the non-CNN models. Besides, we observe an interesting phenomenon that when all the teacher models are selected from CNNs, the attack success rate on the target CNN models are actually not the best. The best results are obtained by selecting 2 CNN and 2 non-CNN models as the teacher models, which implies that learning from both the CNN and non-CNN models is more effective, when attacking the CNN models.

\section{Conclusion}
In this paper, we observe that the output inconsistency problem significantly affects the transferability of adversarial examples. To alleviate this problem while effectively utilizing the existing DNN models, we propose a common knowledge learning (CKL) framework, which distills the knowledge of multiple teacher models with different architectures into a single student model, to obtain better substitute models. Specifically, to emphasize the model-agnostic features, the student model is required to learn the outputs from multiple teacher models. To further reduce the output inconsistencies of models and enhance the adversarial transferability, we propose an input gradient distillation scheme for the student model. Extensive experiments on CIFAR10 and CIFAR100 have demonstrated the superiority of our proposed work.

\bibliographystyle{ACM-Reference-Format}
\bibliography{sample-base}

\clearpage

\appendix
In appendix, we firstly verify that the conflicting gradients do exist when multiple teachers are employed, in Section~\ref{sec:conflicting gradient}. Then, we compare our method with ILA-DA in Section~\ref{sec:ila_da}, to show that our method can also function decently with the intermediate level based attack methods. In Section~\ref{sec:target_c100}, we conduct experiments with the targeted attack settings on CIFAR100. At last, in Section~\ref{sec:ensemble}, we compare our method with ensemble attack, which is a commonly used technique by combining multiple substitute models as the source model in adversarial attack.

\section{Conflicting gradient Phenomenon among Deep Models}
\label{sec:conflicting gradient}
In our CKL method, we adopt PCGrad~\cite{PCGrad} to alleviate the conflicting gradient problem. In this section, we experimentally verify that the conflicting gradient problem does exist. Conflicting gradient is defined by \cite{PCGrad} as Definition 1 presents.

\noindent \textbf{Definition 1.} Two gradients $g_i$ and $g_j$ are called as conflicting when $g_i \cdot g_j < 0$.

\begin{figure}
  \centering
  \includegraphics[width=1\linewidth]{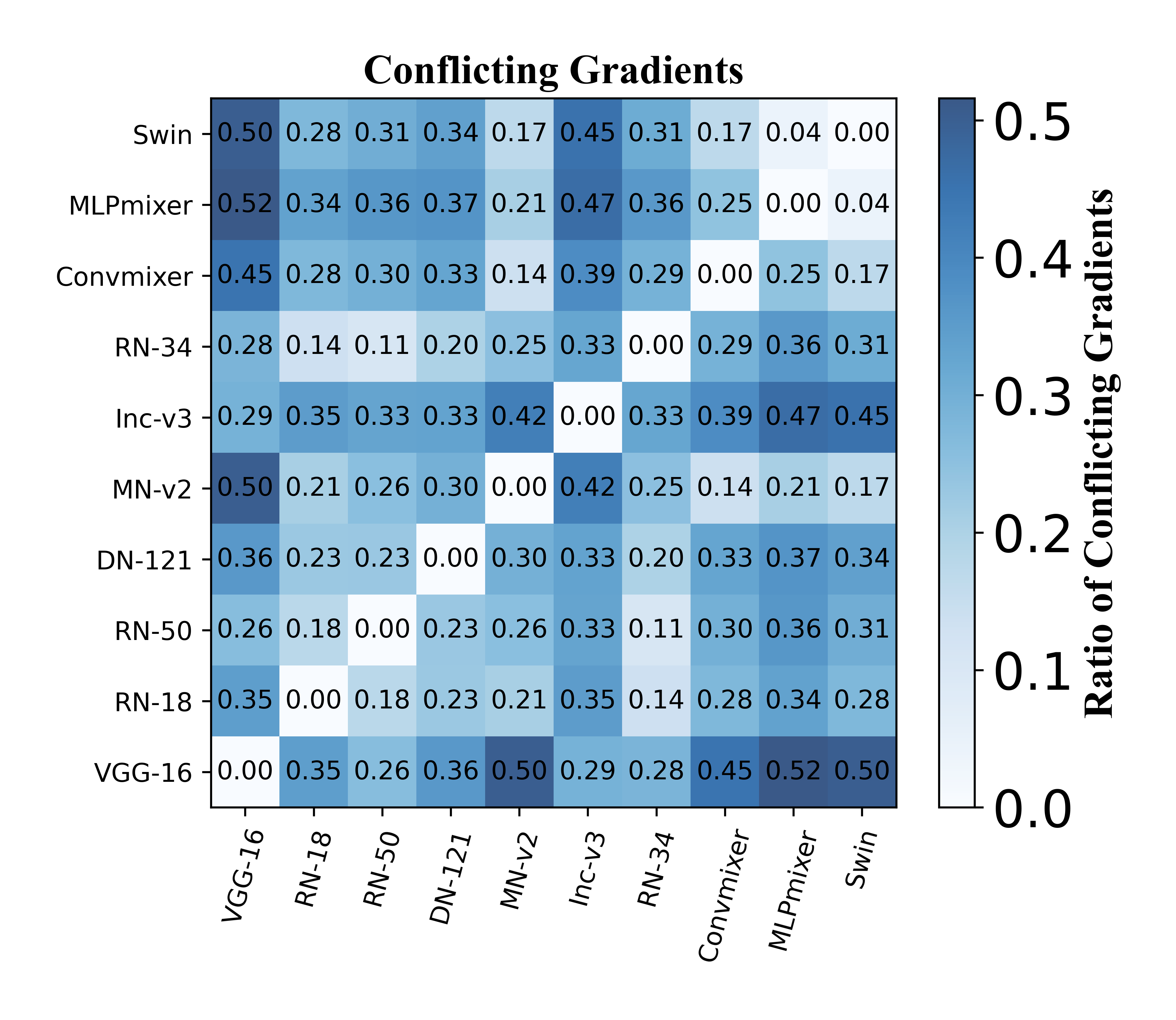}
  \caption{The ratio of conflicting gradients among different models. Higher values indicate more conflicting gradients.}
  \label{fig:conflicting_gradient}
\end{figure}

To verify the that input gradients from different deep models exist conflicts, we employ the CIFAR10 testing set to compute $\nabla_x L_{M_i}(x) \cdot \nabla_x L_{M_j}(x)$, where $x$ denotes the input image, $L_{M_i}(x)$ denotes the loss of the $i$-th model and $\nabla_x L_{M_i}(x)$ is the corresponding input gradient. We define $R_{ij}$ as the ratio of conflicting gradients between $M_i$ and $M_j$, which is formulated as
\begin{equation}\label{eq:ratio_of_conflicts}
  R_{ij} = \frac{\# \{x\in \mathcal{X} | \nabla_x L_{M_i}(x) \cdot \nabla_x L_{M_j}(x)<0 \}}{\# \{x\in \mathcal{X}\}}.
\end{equation}
Here, $\#$ denotes the number of elements in the set and $\mathcal{X}$ is the dataset, e.g., the CIFAR10 testing set in our experiments. We employ VGG-16, ResNet-18, ResNet-50, DenseNet-121, MobileNet-v2, Inception-v3, ResNet-34, Convmixer, MLPmixer and Swin Transformer to conduct the experiment. The results of $R_{ij}$ are shown in Figure~\ref{fig:conflicting_gradient}.

We observe that the conflicting gradient problem is a common phenomenon between different deep models from either different or the same types of DNN architectures. Typically, a higher value indicates that there are more conflicting gradients between the two models. Besides, the ratio of conflicting gradients tends to be higher for two models from two different DNN architectures. Taking VGG-16 as an example, the ratios of conflicting gradients between itself and non-CNN models, i.e., Convmixer, MLPMixer and Swin Transformer, are 0.45, 0.52 and 0.50, respectively, which are usually higher than that between VGG-16 and other CNN models.

\section{Comparison with ILA-DA}
\label{sec:ila_da}
To further demonstrate the effectiveness of our CKL method, we compare it with the SOTA intermediate level based attack method, i.e., ILA-DA~\cite{ila-da}. Since ILA-DA requires a pre-specific intermediate layer to obtain the feature map, it cannot directly employ the Transformer-based models as its source model. Therefore, we respectively utilize VGG-16, ResNet-18 and ResNet-50 as the source model to generate adversarial examples. Target models include CNN and non-CNN models. The results are shown in Table~\ref{tab:ila_da_results}. As can be observed, our CKL method can consistently improve ILA-DA's performances, whatever the target model is. The averaged improvement of the ASR results are more than 18\%, which further proves the effectiveness of our CKL method when being integrated to the intermediate level based attacks.

\begin{table*}[t]
  \caption{Integrating our CKL into SOTA intermediate level attack method on CIFAR10. The first column introduces the source models and the first row presents the test models. We report the averaged attack success rate on the entire testing set. `+CKL' represents that our CKL framework is integrated.}

  \label{tab:ila_da_results}
  \begin{tabular}{c|c|ccccccc|c}
    \toprule
                                 &Attack Method     & ResNet-34   & Inception-v3   & MobileNet-v2     & DenseNet-121    & ConvMixer    & ViT-S      & Swin-T  & Average \\
    \midrule
    \multirow{2}{*} {ResNet-18}  &ILA-DA~\cite{ila-da}            & 69.37        & 59.38         & 70.58            & 66.29           & 55.65        & 33.42      & 43.97   & 56.95 \\
                                 &ILA-DA+CKL        & 82.74        & 79.40         & 85.93            & 84.38           & 79.45        & 59.57      & 80.00   & \pmb{78.78(+21.83)}  \\

    \midrule
    \multirow{2}{*} {ResNet-50}  &ILA-DA            & 70.96        & 67.82         & 78.23            & 76.98           & 62.08        & 35.78      & 45.21   & 62.43  \\
                                 &ILA-DA+CKL        & 82.77        & 84.97         & 90.26            & 86.45           & 84.30        & 58.72      & 81.14   & \pmb{81.23(+18.80}  \\

    \midrule
    \multirow{2}{*} {VGG-16}     &ILA-DA            & 49.69        & 89.02         & 91.66            & 52.59           & 80.04         & 32.34     & 51.42    & 62.97 \\
                                 &ILA-DA+CKL        & 75.80        & 91.33         & 94.95            & 79.81           & 91.15         & 55.53     & 83.33    & \pmb{81.70(+18.73)} \\

    \bottomrule
  \end{tabular}

\end{table*}

\section{Targeted Attack on CIFAR100}
\label{sec:target_c100}
We conduct targeted adversarial attack experiments on the CIFAR100 Dataset. VNI-FGSM and Logit attack are employed as our baselines. We set the maximum perturbation as $\epsilon$=8/255 and maximum number of iterations to 300. The step size is set to 2/255. We report the targeted attack success rate (tASR) in Table~\ref{tab:targeted_c100}. The first column introduces the source models and the first row presents the target models. `Average' represents the average tASR values of all the target models. As can be observed, the tASR scores on CIFAR100 are usually lower than the corresponding results on CIFAR10, as shown in Table 3 of the manuscript, which implies that the targeted attack on CIFAR100 is a more complicated problem. Apparently, our CKL method still significantly ourperforms the baseline methods.

\begin{table*}[t]

  \caption{Targeted attack results on CIFAR100. We generate examples on the testing set and report the tASR value.}

  \label{tab:targeted_c100}
  \begin{tabular}{c|c|cccccc|c}
    \toprule
                                 &Attack Method  & Vgg-16       & Inception-v3  & MobileNet-v2     & DenseNet-121    & ConvMixer    & ViT-S    & Average     \\
    \midrule
    \multirow{4}{*} {ResNet-18}  &VNI~\cite{wang2021enhancing}            & 21.48        & 11.47         & 11.22            & 19.67           & 7.46        & 1.79      & 12.18     \\
                                 &VNI+CKL        & 34.99        & 30.79         & 23.94            & 27.05           & 15.38        & 3.65     & \pmb{22.63(+10.45)}   \\
                                 &Logit~\cite{logit_attack}          & 32.86        & 18.94         & 14.52            & 31.95           & 11.54        & 2.60     & 18.73 \\
                                 &Logit+CKL      & 46.90        & 41.86         & 31.47            & 41.03           & 24.11        & 5.99     & \pmb{31.89(+13.16)}  \\

    \midrule
    \multirow{4}{*} {ResNet-50}  &VNI            & 14.95        & 9.39          & 11.10            & 14.18           & 8.15        & 1.52      & 9.88    \\
                                 &VNI+CKL        & 30.29        & 32.02         & 24.13            & 26.99           & 16.13        & 3.47     & \pmb{22.17(+12.29)}   \\
                                 &Logit          & 23.82        & 16.23         & 14.23            & 27.79           & 12.22        & 2.22      & 16.08\\
                                 &Logit+CKL      & 44.16        & 46.02         & 34.31            & 42.15           & 27.05        & 6.62      & \pmb{33.38(+17.30)}\\

    \midrule
    \multirow{4}{*} {Swin}       &VNI            & 1.16        & 1.01         & 1.58              & 0.78           & 1.53         & 2.29        & 1.39  \\
                                 &VNI+CKL        & 2.50        & 3.41         & 3.48              & 1.61           & 2.70         & 1.82        & \pmb{2.58(+1.19)} \\
                                 &Logit          & 3.30         & 4.07          & 3.74            & 2.62             & 4.42       & 9.02        & 4.52  \\
                                 &Logit+CKL      & 12.39        & 19.87         & 14.55           & 8.87             & 12.48      & 6.07        & \pmb{12.37(+7.85)} \\

    \bottomrule
  \end{tabular}

\end{table*}

\begin{table*}[t]
  \caption{Comparison with ensemble attack on CIFAR10. We report the averaged attack success rate on the entire testing set. `+CKL' represents that our CKL framework is integrated.}

  \label{tab:ensemble_results}
  \begin{tabular}{c|c|cccc|c|c}
    \toprule
    \multicolumn{2}{c|}{ }                      & ResNet-50   & Inception-v3   & Swin-T     & MLPMixer                          &Average of teacher models     \\
    \midrule
    \multicolumn{2}{c|}{Ensemble models}        & 84.14        & \pmb{97.90}  & \pmb{100.0}  &\multicolumn{1}{c|}{\pmb{99.83}}  & \pmb{95.44} \\
    \midrule
    \multirow{3}{*}{CKL}    &ResNet-18          & 86.30        & 83.09         & 82.02       & 71.78                            & 80.79  \\
                            &ResNet-50          & \pmb{89.11}        & 85.69         & 80.12       & 68.64                      & 80.89      \\
                            &Vgg-16             & 79.47        & 91.89         & 84.61       & 68.20                            & 81.04  \\

    \midrule
    \multicolumn{2}{c|}{ }                      & VGG-16    & DenseNet-121   & ConvMixer  & ViT-S                               &Average of unseen models & time (s)\\
    \midrule
    \multicolumn{2}{c|}{Ensemble models}         & 82.66      & 74.11  & 87.82 & 59.79                                          & 76.10        &\multicolumn{1}{|c}{1174.65}\\
    \midrule
    \multirow{3}{*}{CKL}    &ResNet-18          & 87.66        & \pmb{88.37}      & 81.71    & \pmb{62.14}                      & 79.97                   & 41.10\\
                            &ResNet-50          & 86.90        & 87.97      & 83.65      & 58.93                                & 79.36           & 124.30\\
                            &Vgg-16             & \pmb{95.74}        & 84.07      & \pmb{91.54}   & 58.40                       & \pmb{82.43}           & 54.32\\

    \bottomrule
  \end{tabular}

\end{table*}

\section{Comparison with Ensemble attack}
\label{sec:ensemble}
Ensemble attack is a commonly used technique to generate adversarial examples with better transferability, by utilizing multiple source models. On the contrary, our CKL framework distills the knowledge of multiple teacher models into one single student model. Here, we compare our CKL method with the ensemble strategy. MI-FGSM is employed as the attack method. ResNet-50, Inception-v3, Swin-T, and MLPMixer are utilized as the teacher models, which are also the substitute models for the ensemble attack. We conduct this experiment on the CIFAR10 testing set. The ensemble attack is achieved by utilize the averaged value of the four outputs to generate adversarial examples. The results are shown in Table~\ref{tab:ensemble_results}. As can be observed, if the target model is one of the teacher (known) models, ensemble attack gives better performance. Meanwhile, when the target model is unseen, our CKL method obviously outperforms the ensemble strategy, which validates the effectiveness of our common knowledge learning.

Besides, ensemble strategy possesses two obvious defects. Firstly, when there exist non-CNN models in the ensemble models, it cannot employ any intermediate level based attacks, because intermediate level based attacks require a pre-specific intermediate layer to obtain the intermediate level feature map. Unfortunately, the non-CNN models can hardly give feature maps. Secondly, the ensemble strategy tends to induce higher computational complexity during the attack process, especially when the model size of the substitute models are large. Meanwhile, once our student model is obtained, the time cost of our attack process is much lower than the ensemble strategy, because the student model is usually less complex than the teacher models. We compare the running time of generating adversarial examples in the last column of Table~\ref{tab:ensemble_results}, which is obtained on a single RTX 3080Ti GPU. It is obvious that our method is faster than the ensemble attack. When ResNet-18 is employed as the student model, our method (41.10s) is more than $25 \times$ faster than the ensemble strategy (1174.65s).

\end{document}